\documentclass[conference]{IEEEtran}
\IEEEoverridecommandlockouts

\usepackage{cite}
\usepackage{amsmath,amssymb,amsfonts}
\usepackage{algorithmic}
\usepackage{graphicx}
\usepackage{textcomp}
\usepackage{xcolor}
\usepackage{booktabs}
\def\BibTeX{{\rm B\kern-.05em{\sc i\kern-.025em b}\kern-.08em
    T\kern-.1667em\lower.7ex\hbox{E}\kern-.125emX}}
\usepackage{balance}
\usepackage{pifont}
\usepackage[T1]{fontenc}
\usepackage{multirow}
\usepackage{url}
\usepackage{bm}

\def\W{{\mathbf W}}
\def\x{{\mathbf x}}
\def\h{{\mathbf h}}
\def\ii{{\hat{\imath}}}												% Definition of the immaginary symbol i
\def\ij{{\hat{\jmath}}}												% Definition of the immaginary symbol j
\def\ik{{\hat{\kappa}}}												% Definition of the immaginary symbol k

\begin{document}

\title{Towards Explaining Hypercomplex \\ Neural Networks
\thanks{
This work was partially supported by the Italian Ministry of University and Research (MUR) within the PRIN 2022 Program for the project ``EXEGETE: Explainable Generative Deep Learning Methods for Medical Signal and Image Processing", under grant number 2022ENK9LS, CUP C53D23003650001, and in part by the European Union under the National Plan for Complementary Investments to the Italian National Recovery and Resilience Plan (NRRP) of NextGenerationEU,  Project PNC 0000001 D3 4 Health - SPOKE 1 - CUP B53C22006120001.}
}

% \author{\IEEEauthorblockN{Anonymous Authors}%\\
%\IEEEauthorblockA{\textit{Anonymous Affiliations}}

\author{\IEEEauthorblockN{Eleonora Lopez, Eleonora Grassucci, Debora Capriotti and Danilo Comminiello}
        % <-this % stops a space
        \IEEEauthorblockN{\textit{Dept. Information Engineering, Electronics and Telecommunications (DIET), Sapienza University of Rome, Italy}\\
        Email: eleonora.lopez@uniroma1.it.}
}

\maketitle

\begin{abstract}
Hypercomplex neural networks are gaining increasing interest in the deep learning community. The attention directed towards hypercomplex models originates from several aspects, spanning from purely theoretical and mathematical characteristics to the practical advantage of lightweight models over conventional networks, and their unique properties to capture both global and local relations. In particular, a branch of these architectures, parameterized hypercomplex neural networks (PHNNs), has also gained popularity due to their versatility across a multitude of application domains. Nonetheless, only few attempts have been made to explain or interpret their intricacies. In this paper, we propose inherently interpretable PHNNs and quaternion-like networks, thus without the need for any post-hoc method. To achieve this, we define a type of cosine-similarity transform within the parameterized hypercomplex domain. This PHB-cos transform induces weight alignment with relevant input features and allows to reduce the model into a single linear transform, rendering it directly interpretable. In this work, we start to draw insights into how this unique branch of neural models operates. We observe that hypercomplex networks exhibit a tendency to concentrate on the shape around the main object of interest, in addition to the shape of the object itself. We provide a thorough analysis, studying single neurons of different layers and comparing them against how real-valued networks learn. The code of the paper is available at \url{https://github.com/ispamm/HxAI}.
\end{abstract}


\IEEEpeerreviewmaketitle

\section{Introduction}
\label{sec:intro}

\begin{figure}[t]
    \centering
    \includegraphics[width=\linewidth]{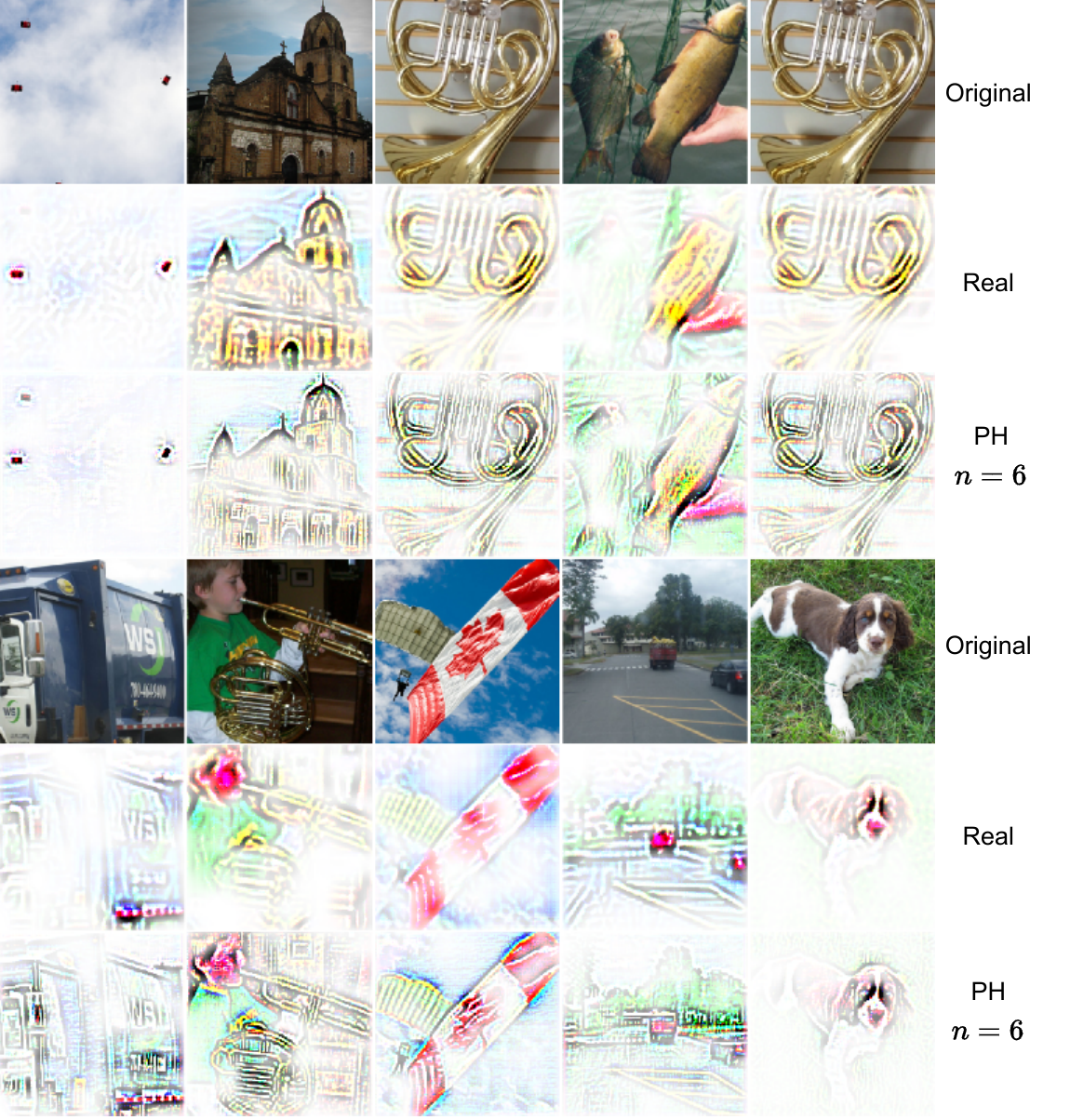}
    \caption{Explanations of standard B-cos networks and PHB-cos networks with $n=6$, with the original sample displayed on top of the maps. The explanations are for the class corresponding to the input image following \eqref{eq:collapsed}.}
    \label{fig:examples}
\end{figure}

The escalating interest in hypercomplex neural networks within the research community has spurred significant exploration into their underlying principles and functionalities. Hypercomplex models attract attention for various reasons. Indeed, many works are studying their theoretical and mathematical attributes. Starting from foundational works on complex-valued networks \cite{hirose2012complex, mandic2009complex}, to the most popular subdomain of hypercomplex numbers, i.e., the quaternion domain \cite{matsui2004quaternion, mandic2010quaternion, ParcolletASRU2017}, and very recent works which study various subdomains and unveil new properties \cite{Zheng2023TCSVT, valle2024universal, Mandic2023TheHE}. These works have shown the unique capacity of these high-dimensional neural networks to model both global and local relations found in multidimensional data \cite{ParcolletAIR2019}. Thanks to this ability, they ensure high performance even with the advantageous parameter reduction that they induce \cite{valle2021hypercomplex, BrignoneTCAS2022}. Moreover, among these hypercomplex architectures, parameterized hypercomplex neural networks (PHNNs) have gained prominence due to their remarkable flexibility across diverse application domains \cite{GrassucciTNNLS2022, PanagosSETN2022, LopezARXIV2023}. They allow the user to define the operating domain of the model by configuring a hyperparameter $n$. For instance, with $n=4$, the PH network functions akin to a quaternion neural network (QNN), while setting $n=3$ enables the processing of RGB images within their natural domain \cite{GrassucciTNNLS2022}.

Despite this surge in interest, there exists a gap in the literature regarding the explainability or interpretability aspects of these intricate models. Only a few attempts have been made in this regard, as evidenced by recent studies \cite{GrassucciTNNLS2022, LopezARXIV2023} that show the application of well-known explainability techniques such as Grad-CAM \cite{selvaraju2017gradcam} and saliency maps \cite{adebayo2018saliency} on PHNNs. However, these works provide only a handful of representative examples without conducting an in-depth investigation, as it was not the primary focus of the respective papers. Therefore, in this paper, we propose \textit{interpretable} PHNNs, thus not requiring any post-hoc explainability method, as we delve into the nuances of these architectures. We achieve this by leveraging the recently proposed B-cos transform \cite{bohle2022bcos} and defining it in the hypercomplex domain. This transform presents two main advantages. Firstly, by optimizing it the weights of the network are pressured to align with relevant patterns in input data, thus becoming highly interpretable. Secondly, it enables the condensation of the entire model into a single linear transform which effectively summarizes the model output \cite{bohle2022bcos}. Therefore, by defining the PHB-cos transform, we bring these properties to PHNNs rendering them inherently interpretable. Furthermore, we extend this approach to quaternion-like models, i.e., a quaternion formulation that we derive from the construction of the weight matrix of PH layers. This extension results in interpretable quaternion-like architectures. To summarize, our contributions are as follows.

\begin{itemize}
    \item We introduce the PHB-cos transform in order to enhance PHNNs and make them interpretable. Furthermore, we formulate this transformation for quaternion-like networks, thereby making them interpretable as well. By doing so, we take a first step towards understanding how this branch of exotic neural networks works and explore how they learn differently compared to standard neural networks.
    \item We demonstrate that the efficacy of the B-cos transform can be translated to these architectures. Specifically, the interpretable PHNN, configured with $n=3$ and $n=6$, retains high performance when trained on Imagenette \cite{howard2020imagenette} and a more challenging dataset, Kvasir \cite{pogorelov2017kvasir}. Similarly, we also evaluate the inherent explanations obtained from the interpretable PHNN. The findings showcase high localization accuracy compared to alternative post-hoc methods, in accordance with the standard B-cos.
    \item We conduct an in-depth analysis of the explanations obtained both from the inherently interpretable PHNN as well as from well-known post-hoc explainability techniques. We draw some interesting insights by comparing them with their respective real-valued counterparts as well as against each other. Moreover, we investigate PHNNs and quaternion-like models down to the single neuron in a specific layer, uncovering intriguing aspects of their learning behavior.
\end{itemize}

The remainder of the paper is organized as follows. Section~\ref{sec:related} provides an overview of related works, encompassing both hypercomplex networks and explainability methodologies. In Section~\ref{sec:method}, we provide a comprehensive background on hypercomplex models and elucidate the proposed method. Section~\ref{sec:results} presents the experimental validation, results and in-depth discussion. Finally, in Section~\ref{sec:conclusion}, we draw the conclusion of this work and pinpoint future directions.

\section{Related works}
\label{sec:related}

\subsection{Hypercomplex neural networks}

In recent years, hypercomplex neural networks have gained increasing attention and several sub-branches have emerged from the Cayley-Dickson algebra systems that comprise complex, quaternion, tessarine, and many other hypercomplex domains. Among them, quaternion numbers are the most widespread due to their properties in building multidimensional representations \cite{grassucci2023grouse, sigillo2023generalizing} of 3D data \cite{BrignoneTCAS2022, Zheng2023TCSVT} and its transformations \cite{QinTPAMI2022} and capturing inter-dimensional relations \cite{Zheng2023TCSVT, ParcolletAIR2019}. Other algebras have been exploited for defining deep learning models with domain-specific properties, such as tessarines that are four-dimensional numbers like quaternions possessing instead product commutativity  \cite{NavarroJFI2020}, and dual quaternions, whose properties include the joint equivariance to rotations and translations in 3D spaces \cite{VieiraMLSP2023} or additional degrees of freedom in describing spatial audio \cite{GRASSUCCIPRL2023}. However, each hypercomplex domain has specific properties that may not hold for different domains. Moreover, hypercomplex algebras rigidly follow the Cayley-Dickson system in which there exists an algebra for each $2^m$ where $m=1,2,\cdots$. Therefore, while more general studies on hypercomplex neural networks have started to appear in the literature \cite{valle2024universal, Parada2023TSP, NittaTSP2019}, a novel method has been developed to overcome domain-specific hypercomplex network limitations by learning algebra rules directly from data. Parameterized hypercomplex neural networks (PHNNs) have been proposed in \cite{GrassucciTNNLS2022, Zhang2021PHM, LeICANN2021} with the breakthrough of learning the algebra structure directly from data. Therefore, PHNNs can exploit hypercomplex advantages for domains for which there does not exist a predefined algebra yet. This feature expands PHNNs effective applicability in several domains of applications including images \cite{GrassucciIJCNN2022}, text \cite{MancanelliMLSP2023}, audio \cite{PanagosSETN2022} and medical data \cite{LopezARXIV2023, Lopez2023ICASSP, lopez2023attention}.

\subsection{Explaining neural networks}

The most popular approach when it comes to explaining neural models is to design a post-hoc method, i.e., a technique applied on a pretrained network in order to interpret its predictions. In turn, within these methodologies, we can distinguish two broad categories: gradient-based and perturbation-based approaches. The first involves computing the gradients of the output with respect to the input via backpropagation to infer attribution scores \cite{nielsen2022robust}. While the latter consist in perturbing input features by altering their values and assessing the impact of these changes on the network performance \cite{nielsen2022robust}. Among the first category, we can find saliency maps \cite{adebayo2018saliency}, Grad-CAM \cite{selvaraju2017gradcam} and attribution propagation methods such as LRP \cite{montavon2019lrp} and CRP \cite{achtibat2023crp}. Instead, LIME is a popular perturbation method \cite{ribeiro2016lime}. Although these techniques are widespread, they are not always reliable as they do not have access to the original training data \cite{laugel2020unjustified}.

Conversely, alternative approaches concentrate on designing architectures that inherently possess interpretability, without the need for any external probe. Generally, they consist in applying some modification of the basic components of a neural network. Among these, there is the filter-based approach \cite{zhang2018interpretable}, the interpretable feedforward design \cite{kuo2019interpretable}, prototype-based networks \cite{chen2019looks}, CoDA-Nets \cite{bohle2021convolutional} and B-cos transform-based networks \cite{bohle2022bcos}. Despite the inherent interpretability offered by these techniques, one drawback is that they are often incompatible with widely used deep learning architectures \cite{bohle2022bcos}, with the exception of the B-cos transform. Indeed, the latter is compatible with a wide range of popular networks, such as ResNets \cite{he2016resnet}, DenseNets \cite{huang2017densenet} and so on \cite{bohle2022bcos}. In this paper, we extend the B-cos transform to the branch of hypercomplex models, in particular PHNNs and quaternion-like models, and prove the efficacy of PHB-cos models both in terms of classification accuracy as well as localization accuracy of the inherent explanation maps.

\section{Proposed method}
\label{sec:method}

In this section, we expound on the proposed method to render PHNNs interpretable. First, we give a detailed overview and background on hypercomplex and PH architectures. Next, we define the PHB-cos transform from which we build hypercomplex B-cos networks that are endowed with inherent interpretability.

\subsection{Hypercomplex neural networks}
\label{sec:hypercomplex}

Hypercomplex neural networks are defined over hypercomplex algebras that can be defined by the Cayley-Dickson construction:

\begin{equation}
\label{eq:cayley}
    n=2^m, \qquad \text{with } m \in \mathbb N.
\end{equation}

From \eqref{eq:cayley} we can derive the most common hypercomplex numbers as complex numbers can be defined for $n=2$, quaternions for $n=4$, octonions for $n=8$, and so on. As it is clear from this formulation, no hypercomplex algebras exist for $n=3, 5, \cdots$, limiting their domain of applications to data with specific dimensions.

\textbf{Quaternions and Hamilton product.} Quaternion numbers are built on four real-valued coefficients and three imaginary units as:

\begin{equation}
    q = q_0 + q_1 \ii + q_2 \ij + q_3 \ik, 
\end{equation}

\noindent in which $\ii^2 = \ij^2 = \ik^2 = -1$ and $\ii \ij = - \ij \ii ; \; \ij \ik = - \ik \ij ; \; \ik \ii = - \ii \ik$. Therefore, the product between two quaternions $q$ and $p$ is not commutative and requires a peculiar definition named Hamilton product:

\begin{equation}
	\begin{split}
	p \times q &= \left(p_{0}q_{0} - p_{1}q_{1} - p_{2}q_{2} - p_{3}q_{3}\right)\\
	&+ \left(p_{0}q_{1} + p_{1}q_{0} + p_{2}q_{3} - p_{3}q_{2}\right)\ii \\
	&+ \left(p_{0}q_{2} - p_{1}q_{3} + p_{2}q_{0} + p_{3}q_{1}\right)\ij \\
	&+ \left(p_{0}q_{3} + p_{1}q_{2} - p_{2}q_{1} + p_{3}q_{0}\right)\ik. \\
    \end{split}
    %e' giusto cosi eh, lo so che non ti sembra haah stavo ragionando se mettere direttamente quello con W
\end{equation}

Quaternion neural networks (QNNs) are defined over the Hamilton product when multiplying a weight matrix $\mathbf{W}$ and the input $\mathbf{x}$ as:

\begin{equation}
\label{eq:qnn}
    \mathbf{y} = \mathbf{W} \times \mathbf{x}.
\end{equation}

\noindent Thanks to the layer formulation in \eqref{eq:qnn}, QNNs allow a $75\%$ reduction of learnable parameters while preserving the performance of the network due to the ability to capture relations among the input dimensions \cite{ParcolletAIR2019}. However, QNNs applicability is limited to 3D or 4D data due to the intrinsic four-dimensional nature of quaternion numbers.

To overcome quaternions limitations while exploiting their advantages with any-dimensional input, parameterized hypercomplex neural networks (PHNNs) have been introduced in \cite{GrassucciTNNLS2022, Zhang2021PHM}. PHNNs can grasp algebra rules directly from data \cite{GrassucciTNNLS2022} thanks to the parameterized hypercomplex convolutional (PHC) layer that is build as sum of Kronecker products between the batch of matrices $\mathbf{A}_i$ that encode the algebra rules and the set of matrices $\mathbf{F}_i$ that represent the filters of the convolution. The layer is parameterized by the hyperparameter $n$ defining the domain in which the model operates by building the weight matrix as:

\begin{equation}
    \W = \sum_{i=1}^n \mathbf{A}_i \otimes \mathbf{F}_i.
\label{eq:phc}
\end{equation}

\noindent By setting $n=2$ the layer emulates the complex domain, with $n=4$ the quaternion one, and so on. It is noteworthy that we can define PHNNs also for $n=3, 5 \cdots$ thanks to the learnable algebra matrices $\mathbf{A}_i$, unlocking the advantage of operating in every data domain with a consequent parameter saving equal to $1/n$.

\subsection{Hypercomplex B-cos networks}
\label{sec:phbcos}

\textbf{PHB-cos networks.} Herein, we present parameterized hypercomplex B-cos networks. These models are composed of PHC-B-cos layers which are defined as follows. Given a layer $l$ of a PHB-cos network $f_{\text{PHB-cos}}(\x) = l_N \circ l_{N-1} \circ \dots \circ l_2 \circ l_1(\x)$, the PHC-B-cos layer is

\begin{equation}
    \text{PHC-B-cos}_l(\h, \W) = (\widehat{\W} * \h) \odot |\text{cos}(\h, \widehat{\W})|^{\text{B}-1},
    % ||\mathbf{\hat{w}}|| ||\mathbf{x}|| |c(\mathbf{x}, \mathbf{\hat{w}})|^B 
\label{eq:phcbcos}
\end{equation}

\noindent where $\h$ is the input to layer $l$ and the weight matrix $\widehat{\W}$ is constructed as a sum of Kronecker products according to \eqref{eq:phc}. In addition, its rows $\hat{\mathbf{w}}$ are normalized such that $||\hat{\mathbf{w}}||=1$. The cos function computes the cosine of the angle between the rows of the weight matrix $\widehat{\W}$ and the input $\h$. The power, absolute value, and the $\odot$ operator are applied element-wise. Indeed, the convolution operator is also known as a sliding dot product, thus in the end the operation for a single row $\hat{\mathbf{w}}$ and $\h$ becomes 

% \begin{align*}
%     \text{PHB-cos}(\x, \mathbf{w}) &= ||\mathbf{\hat{w}}|| ||\mathbf{x}|| \text{cos}(\mathbf{x}, \mathbf{\hat{w}}) |\text{cos}(\mathbf{x}, \mathbf{\hat{w}})|^{\text{B}-1}
%     & = ||\mathbf{\hat{w}}|| ||\mathbf{x}|| |\text{cos}(\mathbf{x}, \mathbf{\hat{w}})|^{\text{B}} \times sgn(\text{cos}(\mathbf{x}, \mathbf{\hat{w}})).
% \label{eq:phbcos}
% \end{align*}

\begin{equation}
\begin{split}
    \text{PHB-cos}(\h, \mathbf{w}) &= ||\mathbf{\hat{w}}|| ||\h|| \text{cos}(\h, \mathbf{\hat{w}}) |\text{cos}(\h, \mathbf{\hat{w}})|^{\text{B}-1} \\
    &= ||\mathbf{\hat{w}}|| ||\h|| |\text{cos}(\h, \mathbf{\hat{w}})|^{\text{B}} \text{sgn}(\text{cos}(\h, \mathbf{\hat{w}})).
\end{split}
\label{eq:phbcos}
\end{equation}

\noindent During the optimization process, the objective is to maximize the network output. This entails maximizing the cosine similarity between the weights and input, and aligning the weights closely with the input data. In fact, at the limit, $\hat{\mathbf{w}}$ and $\mathbf{h}$ share a similarity of 1, and thus \eqref{eq:phbcos} becomes exactly equal to $||\h||$. Therefore, this formulation implies a bound on the output of a PHC-B-cos layer equal to $||\h||$. Consequently, the weights achieve a high level of interpretability as they are required to align with the input, thus capturing task-relevant input patterns.

Furthermore, since $f_{\text{PHB-cos}}$ is composed of many layers of PHC-B-cos defined in \eqref{eq:phcbcos}, which in the end applies a linear transform \eqref{eq:phbcos}, a sequence of these layers can be collapsed into one in the following manner. Given a layer $l$ as defined in \eqref{eq:phcbcos} we can rewrite it as

\begin{equation}
    \text{PHC-B-cos}_l(\h, \W) = \mathbf{H}(\h) \h
\end{equation}

\noindent where $\mathbf{H}(\h) = |\text{cos}(\h, \widehat{\W})|^{\text{B}-1} \odot \widehat{\W}$ and $\odot$ scales the rows of the weight matrix $\widehat{\W}$ by the scalar entries of the vector $|\text{cos}(\h, \widehat{\W})|^{\text{B}-1}$. Then, $f_{\text{PHB-cos}}(\x)$ can be written as

\begin{equation}
\begin{split}
    f_{\text{PHB-cos}}(\x) &= \mathbf{H}_N (\h_N) \mathbf{H}_{N-1}(\h_{N-1}) \dots \mathbf{H}_1(\x)\x \\
    &=\prod_{i=1}^{N}\mathbf{H}_i(\h_i) \x \\
    &=\mathbf{H}_{1 \rightarrow N}(\x).
\end{split}
\label{eq:collapsed}
\end{equation}

\noindent Thus the whole network is effectively summarized by a single linear transform $\mathbf{H}_{1 \rightarrow N}(\x)$. Therefore, the inherent explanation of a specific class is simply the row of $\mathbf{H}_{1 \rightarrow N}(\x)$ corresponding to the class, which is what we display in Figs.~\ref{fig:examples} and \ref{fig:kvasir}. Moreover, we can also visualize the contributions of the input $\x$ to a single neuron $i$ in layer $l$, which is $c_i^l(\x)=\mathbf{H}_{1 \rightarrow l}(\x)_i^T \x$, i.e., by taking the $i$-th row up to layer $l$ and then multiply it by the input. Then, the contribution of a single input pixel $(x, y)$ is given by $\sum_{j=1}^{\#\text{ch}}c_i^l(\x)_{(x, y, j)}$, where \#ch is the number of channels of the input \cite{bohle2022bcos}.

With this definition of PHB-cos networks, we maintain the advantages of PH models described in Section~\ref{sec:hypercomplex} which allow the network to operate in a user-defined hypercomplex domain controlled by the hyperparameter $n$, while also bringing interpretability to this branch of networks. To conclude, also PHM-B-cos layers could be employed depending on the need. In this case, we take as base architecture a DenseNet which is fully convolutional, therefore we did not need PHM-B-cos layers.

\textbf{Quaternion-like B-cos networks.} We can similarly extend this concept to quaternion-like models as well. We refer to quaternion-like networks as a quaternion formulation derived from \eqref{eq:phc}. Thus, the weight matrix of a specific layer is constructed from \eqref{eq:phc} with $n=4$ and matrices $\mathbf{A}_i$ fixed, following the rules of the Hamilton product:

\begin{equation}
\begin{aligned}
{\mathbf{A}_1} = \begin{bmatrix}
    1 & 0 & 0 & 0 \\
    0 & 1 & 0 & 0 \\
    0 & 0 & 1 & 0 \\
    0 & 0 & 0 & 1
\end{bmatrix}, \;
{\mathbf{A}_2} &= \begin{bmatrix}
    0 & -1 & 0 & 0 \\
    1 & 0 & 0 & 0 \\
    0 & 0 & 0 & -1 \\
    0 & 0 & 1 & 0
\end{bmatrix}, \\
{\mathbf{A}_3} = \begin{bmatrix}
    0 & 0 & -1 & 0 \\
    0 & 0 & 0 & 1 \\
    1 & 0 & 0 & 0 \\
    0 & -1 & 0 & 0
\end{bmatrix}, \;
{\mathbf{A}_4} &= \begin{bmatrix}
    0 & 0 & 0 & -1 \\
    0 & 0 & -1 & 0 \\
    0 & 1 & 0 & 0 \\
    1 & 0 & 0 & 0
\end{bmatrix}.
\end{aligned}
\label{eq:matrices}
\end{equation}

\noindent It follows that we obtain a quaternion-like weight matrix:

\begin{equation}
{\bf{W}} = \left[ {\begin{array}{*{20}c}
   \hfill {{\bf{W}}_0 } & \hfill { - {\bf{W}}_1 } & \hfill { - {\bf{W}}_2 } & \hfill { - {\bf{W}}_3 } \\
   \hfill {{\bf{W}}_1 } & \hfill {{\bf{W}}_0 } & \hfill { - {\bf{W}}_3 } & \hfill {{\bf{W}}_2 } \\
   \hfill {{\bf{W}}_2 } & \hfill {{\bf{W}}_3 } & \hfill {{\bf{W}}_0 } & \hfill { - {\bf{W}}_1 } \\
   \hfill {{\bf{W}}_3 } & \hfill { - {\bf{W}}_2 } & \hfill {{\bf{W}}_1 } & \hfill {{\bf{W}}_0 } \\
\end{array}} \right].
\label{eq:quat_matrix}
\end{equation}

\noindent Having constructed the weight matrix in this manner, we can directly apply the definition of PHC-B-cos \eqref{eq:phcbcos} and accordingly also \eqref{eq:phbcos}. In other words, we just construct the weight matrix in a different way, thus all the considerations that follow apply herein as well. %However, this is only an approximation of an authentic quaternion network, in which more geometrical considerations need to be taken into account to formulate the quaternionic B-cos transform.

\section{Experiments and results}
\label{sec:results}

\subsection{Dataset and preprocessing}
\label{sec:data}

We employ two datasets for our experiments. The first is the Imagenette dataset \cite{howard2020imagenette}. It comprises a subset of the ImageNet dataset with 10 classes. %and provides images with the full resolution of $375\times500$, as well as train/test splits. 
The second dataset is Kvasir, a dataset for computer-aided gastrointestinal disease detection. It provides RGB images with 8 categories each with 500 samples. %We create train and test splits by taking 80\% of the samples for training and the rest for testing in a stratified fashion.
For both datasets, we apply the same preprocessing and data augmentation. Initially, given an image with three RGB channels, other three channels are added such that the final preprocessed image has six channels $[r,g,b,1-r,1-g,1-b]$, following \cite{bohle2022bcos}. This allows to directly decode the colors from the explanation given by \eqref{eq:collapsed} \cite{bohle2022bcos}. For quaternion-like models we add two additional padding channels. Then, for the training sets, we apply the RandAugment augmentation strategy \cite{cubuk2020randaugment} choosing two random transforms each time, then we apply a random resized crop of $224\times224$, random horizontal flip, color jitter, and lighting noise. While test images are just resized to $256\times256$ and center cropped at the same resolution of training.

\subsection{Architectural details}

We employ as base architecture a DenseNet121 \cite{huang2017densenet}. Thus, for the PH version we substitute convolutional layers with PHC layers, resulting in PHDenseNet121, while for the PHB-cos version, we substitute them with PHC-B-cos layers defined in~\eqref{eq:phcbcos}. For PH models, we test two hypercomplex domains by setting the hyperparameter $n=3$, which allows to process RGB images in their natural domain, and $n=6$ to further exploit the additional channels added during preprocessing as described in Section~\ref{sec:data} (only for B-cos models). %We change the number of filters such that they are divisible by $n$ \cite{GrassucciTNNLS2022}. The weights are initialized with a standard Xavier scheme for experiments on Imagenette, while for experiments on Kvasir they follow the initialization scheme proposed in \cite{GrassucciIJCNN2022}. 
For B-cos networks, both in the real-valued and PH case, we apply the same main configuration suggested in the original paper \cite{bohle2022bcos}. Thus, we set the hyperparameter $\text{B}=2$ and add MaxOut non-linearities with 2 units \cite{goodfellow2013maxout}. Accordingly, we remove batch normalization layers and other non-linearities, such as ReLU and pooling operations.

\subsection{Training recipe}
All experiments follow the same training recipe proposed for B-cos DenseNet121 in the original paper \cite{bohle2022bcos}, both for baselines (standard non-interpretable networks) and B-cos models. The models are trained with an initial learning rate of $10^{-5}$ with a warm-up for 10 epochs and a cosine scheduler, for a total of 200 epochs. We employ the binary cross entropy (BCE) loss and we one-hot encode the target labels with the Adam optimizer and a batch size of 128.

\subsection{Results}

In this section, we provide a comprehensive overview of our experimental validation and corresponding analyses. We begin by presenting a performance analysis of PHB-cos networks. Subsequently, we delve deep into the single neurons of the networks performing a comparative assessment of real-valued and various configurations of PH models. Finally, we verify the efficacy of B-cos explanations when integrated into PH architectures, both from a qualitative and quantitative point of view with respect to established post-hoc explainability methods.

\begin{figure*}[t]
    \centering
    \includegraphics[width=\textwidth]{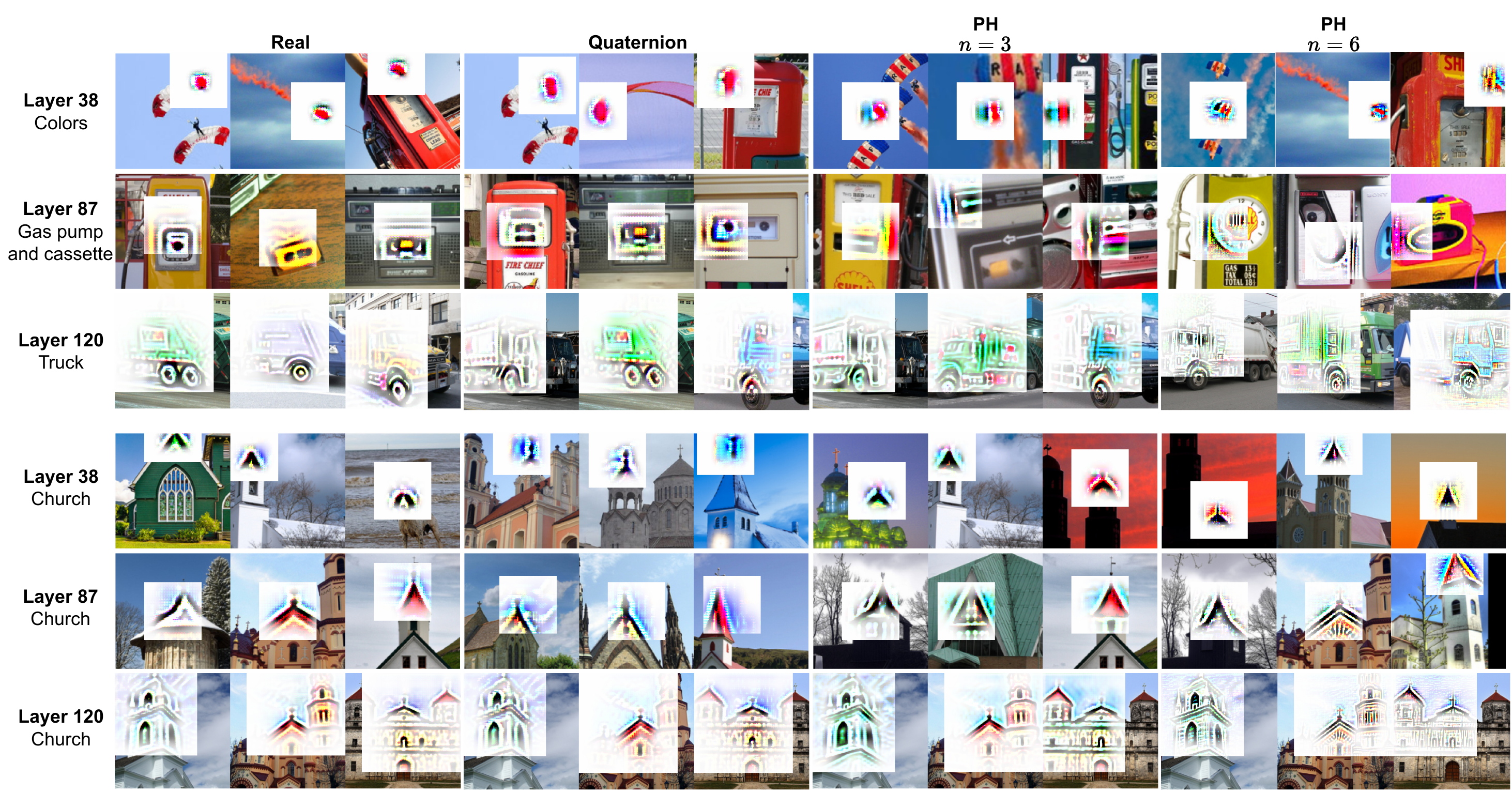}
    \caption{Explanations of single neurons from different layers, displaying a variety of encoded concepts. In detail, for each neuron, we display three images among those contributing to the highest activations. For every image, in correspondence to the highest activating portion, the associated explanation is depicted.}
    \label{fig:neurons}
\end{figure*}

\subsubsection{Model performance}

\begin{table}[t]
\centering
\caption{Results of baselines models (top) and interpretable networks (bottom). Values in bold and underlined represent the best and second-best accuracies, respectively. %Parameterized hypercomplex networks outperform each respective counterpart. PHBcosDenseNet121 with $n=6$ is the most performing model among interpretable ones, with only 1.6M parameters.
}
\begin{tabular}{llccc}
\toprule
\multicolumn{1}{l}{} & \multicolumn{1}{l}{Model} & \multicolumn{1}{c}{Params} & \multicolumn{1}{c}{Imagenette} & \multicolumn{1}{c}{Kvasir} \\ \midrule 
\multirow{3}{*}{Baselines} & DenseNet121 & 5.7M & \underline{94.06} & \textbf{90.75}\\
& Quaternion DenseNet121 & 1.8M & 93.86 & 88.13\\
& PHDenseNet121 ($n=3$) & 2.5M & \textbf{94.52} & \underline{90.63}\\
\midrule
\multirow{4}{*}{B-cos} & DenseNet121 & 11.3M & 92.94 & \underline{87.38}\\
& Quaternion DenseNet121 & 2.7M & 92.41 & 84.38 \\
& PHDenseNet121 ($n=3$) & 3.8M & \underline{92.99} & \textbf{87.75}\\
& PHDenseNet121 ($n=6$) & 1.6M & \textbf{93.15} & 83.75\\
\bottomrule
\end{tabular}
\label{tab:res}
\end{table}

To assess model performance, we employ a comprehensive set of baselines and interpretable networks, with the corresponding results presented in Tab.~\ref{tab:res}. As non-interpretable baselines, we include a real-valued DenseNet121, a PHDenseNet121 with $n=3$, and a Quaternion DenseNet121. In contrast, regarding interpretable models, we train a standard B-cos DenseNet121, two PHDenseNet121 with $n=3$ and with $n=6$, and a quaternion-like DenseNet121 (which for convenience we refer to as ``quaternion" in the table). We set $n=3$ since the images are RGB and we additionally test $n=6$ accounting for the 3 additional channels added during preprocessing for B-cos networks. We train all the mentioned architectures on two different datasets, Imagenette and Kvasir. In accordance with the original paper that introduced B-cos architectures, we can see that there is a slight trade-off between model performance and interpretability. Nonetheless, we demonstrate the efficacy of PHB-cos networks, which indeed maintain the advantages of PH models, thus outperforming the respective real-valued counterparts. In particular, for Imagenette, the configuration with $n=6$ yields the highest accuracy of 93\% with the $1/6$ of parameters, i.e. only 1.6 million parameters. Indeed, it effectively models the 6 channels provided in input as shown in similar works \cite{GrassucciTNNLS2022, LopezARXIV2023, lopez2023attention}. Instead, being Kvasir a more challenging dataset, it appears that the configuration with $n=6$ is undersized as it struggles to reach the performance of its equivalent with $n=3$. This pattern is consistent across the baselines, where the PH model falls slightly short of its real-valued counterpart.

\subsubsection{Qualitative analysis}

\begin{figure}[t]
    \centering
    \includegraphics[width=0.93\linewidth]{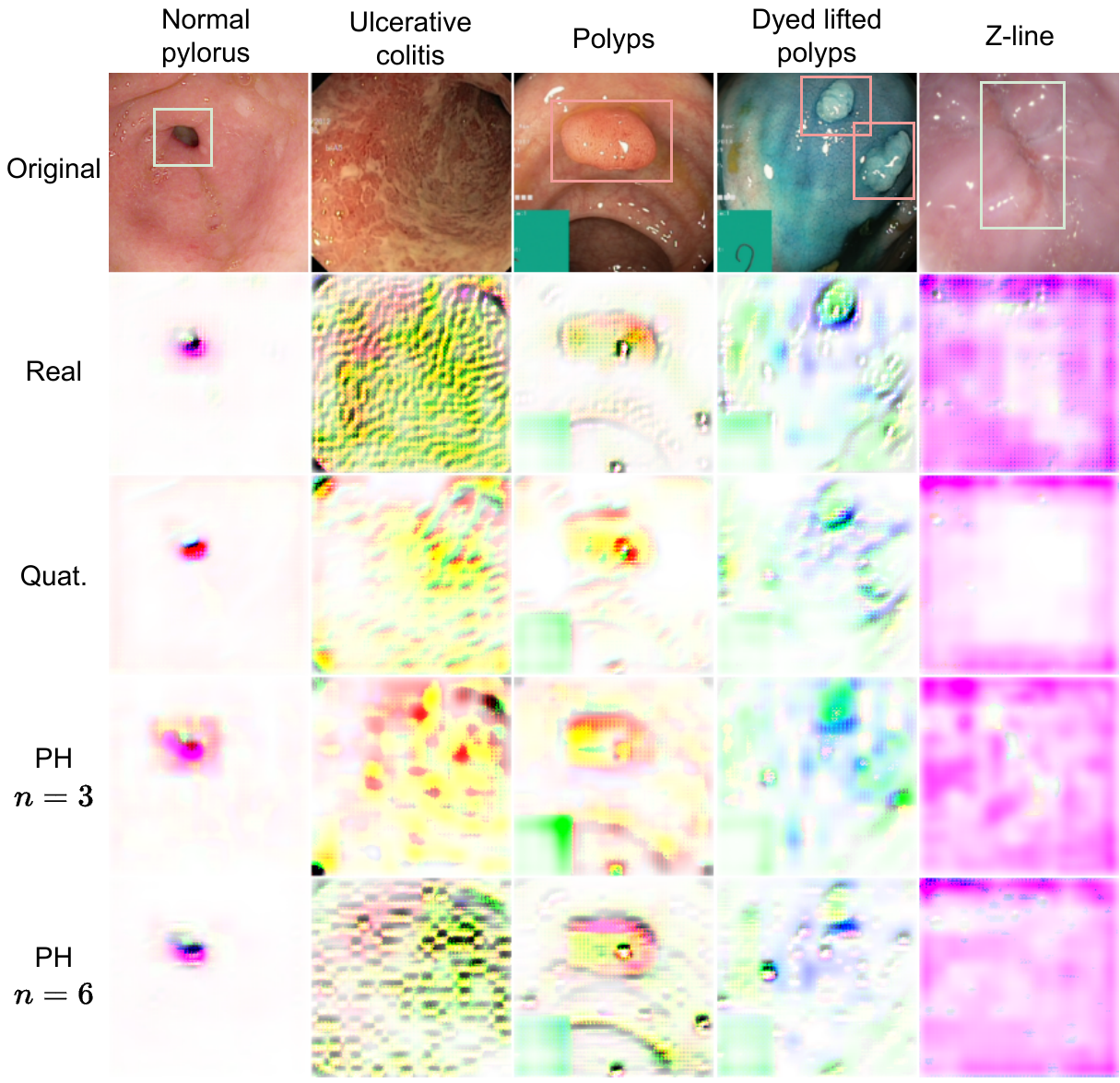}
    \caption{Input images of Kvasir dataset corresponding to different categories (top row) and corresponding class explanations given by the different tested architectures (other rows). Bounding boxes indicate the ROI.}
    \label{fig:kvasir}
\end{figure}

\begin{figure}[t]
    \centering
    \includegraphics[width=\linewidth]{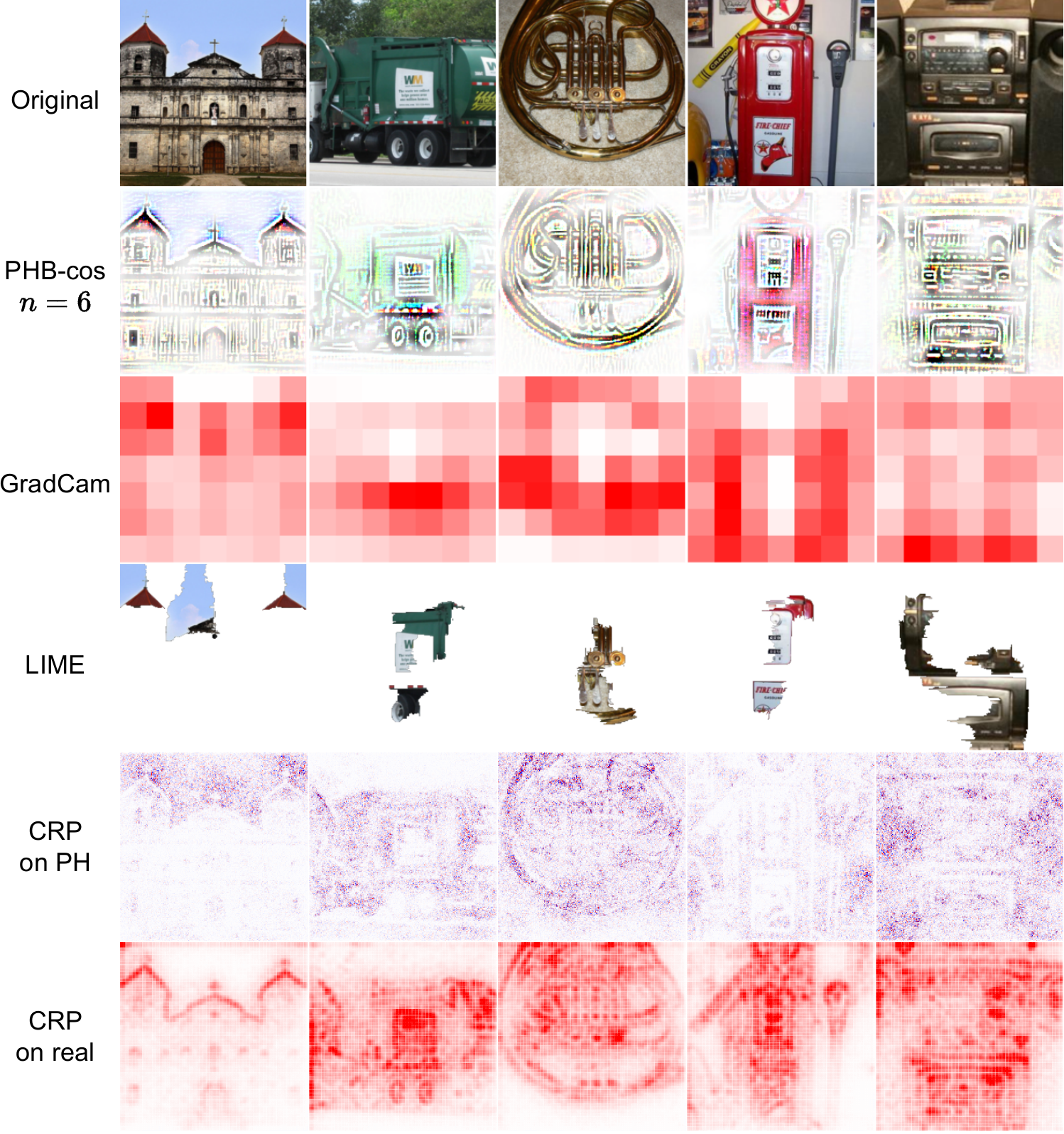}
    \caption{Original input images of Imagenette (top row) and explanations given by the PHB-cos inherent method as well as post-hoc methods applied on PHB-cos with $n=6$, except for CRP which is applied on a PHResNet50 with $n=3$ and a ResNet50 (real) trained on ImageNet.}
    \label{fig:post-hoc}
\end{figure}

This section provides a thorough qualitative exploration of model explanations. Firstly, Figure~\ref{fig:examples} presents a comparison of explanations between a real-valued model and our best PH model. Then, Figure~\ref{fig:neurons} delves into the intricate details of single layers and neurons, showcasing nuanced insights into different levels of the networks and comparing each trained model presented in Tab.~\ref{tab:res}. Subsequently, we display the explanations for the Kvasir dataset for different non-pathological and pathological findings in Fig.~\ref{fig:kvasir}. Finally, we compare the inherent explanations of our best interpretable model PHB-cos with $n=6$ against three well-established post-hoc explainability methods, i.e. GradCam \cite{selvaraju2017gradcam}, LIME \cite{ribeiro2016lime}, and  CRP \cite{achtibat2023crp} in Fig~\ref{fig:post-hoc}. For all figures, we follow the same visualization details as in \cite{bohle2022bcos}.

\textbf{Imagenette results and neuron analysis.} To begin with, what becomes apparent right away from Fig.~\ref{fig:examples} is the fact that explanations of the PH network with $n=6$ are much sharper and additionally grasp more details. For example, in the first image of a parachute, all three parachutes can be seen in the explanation of the latter, while the real model sees only two. Similarly, the PH model captures both fish in the fourth sample, unlike the real counterpart. But perhaps what is most interesting is the sharpness of the explanations, which is a unique aspect of the configuration with $n=6$. In fact, this is apparent from Fig.~\ref{fig:neurons}, where we can see that the sharpness is present only for this architecture and not for the PH model with $n=3$ or the quaternion-like network.

In Fig.~\ref{fig:neurons} we show different concepts learned by specific neurons at different depths of the neural models, that is at layers 38, 87, and 120 (that is the last convolutional layer before the classification layer) of a DenseNet121. The top portion of the figure, displays different concepts for each of these three layers, while the bottom portion shows how the same concept is modeled at different levels of depth of the architectures. The white area corresponds to the highest activations and in that area, we illustrate the respective explanation. First off, from these examples, it is evident that similarly to real-valued networks, PH models capture low-level concepts at early layers (e.g., colors or edges in layer 38), while deeper layers 87 and 120 grasp higher-level concepts (e.g., the features of a cassette or roofs and windows of churches) as expected. More interesting is how these networks learn and what differences are there with respect to real counterparts as well as nuanced variations among different configurations of PH models. For example, it is intriguing that PH networks seem to learn not only the shapes of objects but also focus significantly on the shape \textit{around} the object. This unique characteristic is visible starting from early layers. For example, in the first row, we show how the different networks encode the concept of the color red. In the explanations corresponding to the real model, we can see that the concept of red is indeed a small red portion of the image. Instead, for PH networks, the color red appears \textit{next to} another color, in this case blue, meaning that it focuses also on the surroundings. This is more visible for the version with $n=3$, while the equivalent with $n=6$ seems to encode even more colors at once. In contrast, the quaternion-like network is most similar to the real one. In the second row, we display the neurons that learn concepts of gas pump and cassette at layer 87. Similarly, the real and quaternion networks seem to focus on a central portion of the gas pump/cassette, while PH networks more on the area around the boundary of the object. Finally, in the third row, we illustrate the notion of ``truck" at layer 120. We can observe that the real-valued network puts a lot of emphasis on the tires, while the PH models focus also on the structure of the truck itself and the quaternion network falls in between. We hypothesize that this might be due to the learning behavior of hypercomplex models described above. By focusing on the surroundings around the most informative portion of the input, i.e., the tire, the hypercomplex networks also capture meaningful information about the rest of the truck. Finally, regarding the bottom of the picture displaying the concept of ``church", we can notice that starting from early layers the model operating with $n=6$ is already grasping more details than the one with $n=3$, and both of them more than the real-valued counterpart. Also in later layers such as layer 87, we can observe the amount of details on the middle sample of the $n=6$ configuration with respect to the same input sample of the real model, as well as in layer 120.

\textbf{Kvasir results.} In Fig.~\ref{fig:kvasir} we show different samples falling into non-pathological and pathological categories. The former includes normal pylorus, which is the area around the opening from the stomach into the first part of the small bowel, and the z-line, which marks the transition site between the esophagus and the stomach, both important to be detected \cite{pogorelov2017kvasir}. The latter includes ulcerative colitis, a chronic inflammatory disease affecting the large bowel, polyps, lesions within the bowel detectable as mucosal outgrows, and dyed lifted polyps, which is a polyp lifted by injection and its detection is crucial for polyp removal \cite{pogorelov2017kvasir}. In this case, the most performing model is the hypercomplex network with $n=3$. Indeed, we can observe quite a difference between the explanations of this model and the other networks, even the hypercomplex ones. The colors learned appear very different, this is most evident in the sample of ulcerative colitis for which all the other models encode the color green. Instead, the architecture with $n=3$ encodes red and yellow, thus being more faithful to the original input sample in which the mucosa appears swollen and red. Moreover, the same behavior of hypercomplex models described in the previous section can be observed herein as well, more evidently for the network operating with $n=3$. For example, by observing the explanations of the first sample of this model, a highlighted area in pink around the region of interest (ROI) is clearly visible. Thus, again suggesting that PH models rely also on the shapes \textit{around} the ROI.

\textbf{Comparison to post-hoc.} In Fig.~\ref{fig:post-hoc} we illustrate a comparison with other popular post-hoc methods applied to PHB-cos with $n=6$, except for the CRP strategy. The latter is applied on another model, i.e., a pretrained PHResNet50 with $n=3$ on ImageNet. That is because utilizing CRP with a DenseNete architecture requires a specific implementation. Moreover, being our models trained on a subset of ImageNet, the results are comparable. We additionally show the results of CRP applied on the pretrained real-valued counterpart ResNet50, as the contrast is intriguing. With these comparisons, we verify the qualitative clearness of B-cos explanations with respect to other techniques when integrated inside hypercomplex layers. More interestingly, also from CRP explanations of the PH network, we can observe how the \textit{areas surrounding} the object of interest, e.g., the points of the roofs of the church, are accentuated. This is in contrast with the attribution maps of its counterpart, where the shape of the object itself is outlined, rather than its surroundings. This further emphasizes the learning behavior that has been discussed in previous sections.

\begin{figure}[t]
    \centering
    \includegraphics[width=0.85\linewidth]{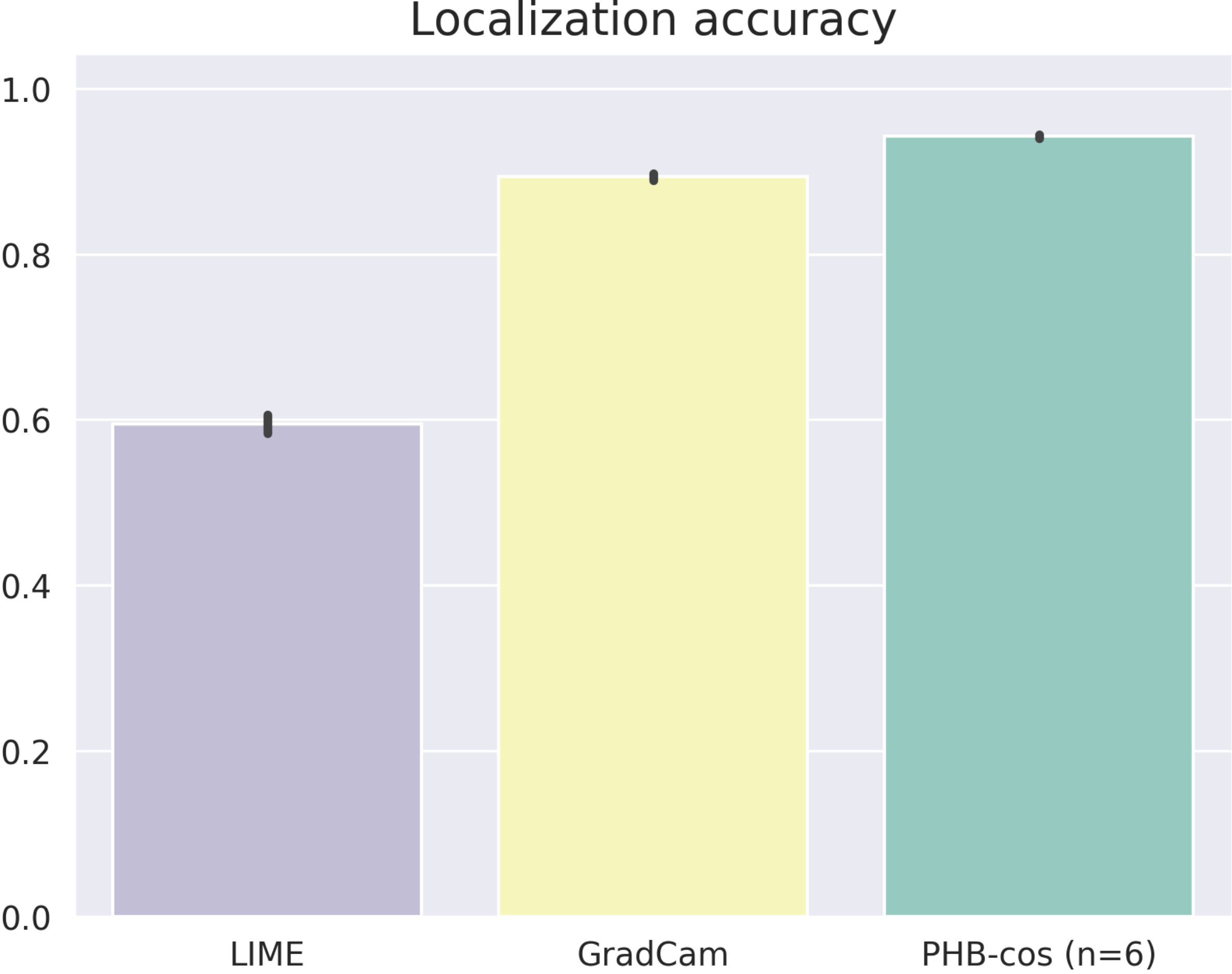}
    \caption{Localization accuracy computed with the grid pointing game for inherent explanations given by PHB-cos with $n=6$ as well as post-hoc methods LIME and GradCam.}
    \label{fig:localization}
\end{figure}

\subsubsection{Quantitative analysis}

We demonstrate the quantitative effectiveness of PHB-cos explanations by employing the same metric used in the paper that introduced the B-cos transform, i.e., the localization accuracy. This is computed through a grid pointing game \cite{bohle2021convolutional}. It involves assessing the trained models on a constructed grid, in our case $2\times2$, containing images from various classes. For each set of class logits, the extent to which an explanation method assigns positive attribution to the correct location within the grid is measured. In accordance with \cite{bohle2021convolutional}, we generate 500 image grids using the most confidently and accurately classified images. We evaluate the PHB-cos explanations for the model with $n=6$ and the explanations of post-hoc methods, except CRP since it evaluates a different network. The results are plotted in Fig.~\ref{fig:localization} and align with the findings of the standard B-cos transform also when integrated within hypercomplex networks. It is evident that the explanations of the PHB-cos network are much more faithful compared to the other two techniques.

\section{Conclusion}
\label{sec:conclusion}

This paper introduces inherently interpretable PHNNs and quaternion-like networks by defining the PHB-cos transform, enabling weight alignment and direct interpretability without post-hoc methods. We conduct a thorough analysis, evaluating the efficacy of the B-cos transform in a hypercomplex domain and qualitatively investigate the explanations. Our study reveals that PH networks tend to focus on both the shape \textit{around} the main object of interest and the object shape itself. Future work aims to investigate and explain how hypercomplex networks learn both global and local relations, as well as generalize this work to multi-view architectures.

\bibliographystyle{IEEEtran}
\bibliography{xPHBiblio}

\end{document}